\pdfoutput=1

\documentclass[11pt]{article}

\usepackage[preprint]{config/acl}

\usepackage{times}
\usepackage{latexsym}

\usepackage[T1]{fontenc}

\usepackage[utf8]{inputenc}

\usepackage{microtype}

\usepackage{inconsolata}

\usepackage{url}            
\usepackage{booktabs}       
\usepackage{nicefrac}       
\usepackage{xcolor}         
\usepackage{multirow}
\usepackage{bm}

\usepackage{amsmath}

\usepackage[ruled]{algorithm2e}
\usepackage{algpseudocode}
\usepackage{graphicx}  
\usepackage{CJK}
\usepackage{bbm}
\usepackage{float}
\usepackage{xcolor,colortbl}

\usepackage{bigstrut,bigdelim}
\usepackage{diagbox}
\usepackage{wrapfig}
\usepackage{verbatim}
\usepackage{afterpage}
\usepackage{enumitem}

\usepackage{subcaption}
\usepackage{multicol}
\usepackage{multirow}
\usepackage{amssymb}
\usepackage{xcolor}
\usepackage{caption}
\usepackage[capitalize]{cleveref}
\usepackage{acronym}
\usepackage{xspace}
\usepackage{pifont}
\usepackage{lipsum}
\usepackage{makecell}
\usepackage{tablefootnote}
\usepackage{footnote}
\usepackage[normalem]{ulem}

\usepackage{listings}
\usepackage{titletoc}

\makeatletter
\DeclareRobustCommand\onedot{\futurelet\@let@token\@onedot}
\def\@onedot{\ifx\@let@token.\else.\null\fi\xspace}
\def\eg{\emph{e.g}\onedot} 

\def\ie{\emph{i.e}\onedot}

\def\etc{\emph{etc}\onedot}

\makeatother

\makeatletter
\renewcommand{\paragraph}{%
  \@startsection{paragraph}{4}%
  {\z@}{0ex \@plus 0ex \@minus 0ex}{-1em}%
  {\hskip\parindent\normalfont\normalsize\bfseries}%
}
\makeatother

\makeatletter
\newcommand{\thickhline}{%
    \noalign {\ifnum 0=`}\fi \hrule height 1pt
    \futurelet \reserved@a \@xhline
}
\makeatother

\DeclareMathOperator*{\argmax}{arg\,max}

\newcommand{\one}{\ding{202}\xspace}
\newcommand{\two}{\ding{203}\xspace}
\newcommand{\three}{\ding{204}\xspace}

\newcommand\DoToC{%
  \startcontents
  \printcontents{}{1}{\textbf{\Large Table of Contents}\vskip5pt\hrule\vskip3pt}
  \vskip3pt\hrule\vskip5pt
}

\acrodef{nlp}[NLP]{natural language processing}
\acrodef{vqa}[VQA]{Visual Question Answering}
\acrodef{cs}[CS]{coherence scoring}
\acrodef{avsd}[AVSD]{coherence scoring}
\acrodef{roi}[RoI]{Region-of-Interest}
\acrodef{vlp}[VLP]{video-language pretrained model}
\acrodef{vlm}[VLM]{vision-language model}
\acrodef{plm}[PLM]{pre-trained language model}
\acrodef{videoqa}[VideoQA]{video question answering}
\acrodef{of}[OF]{optical flow}
\acrodef{llm}[LLM]{large language model}

\usepackage{ulem}

\newcommand{\prompt}[1]{{\fontfamily{lmtt}\selectfont #1}\xspace}
\newcommand{\model}{TGB\xspace}

\newcommand{\longname}{Temporal Grounding Bridge\xspace}

\title{Efficient Temporal Extrapolation of Multimodal Large Language Models with Temporal Grounding Bridge}

  \author{Yuxuan Wang$^{1,3}$ , Yueqian Wang$^{2}$ , Pengfei Wu$^{2}$ \\ \textbf{Jianxin Liang$^{2}$, Dongyan Zhao$^{2,3}$, Yang Liu$^{2,3}$, Zilong Zheng$^{1,3,}$\thanks{Corresponding author.}} \\
$^1$ Beijing Institute for General Artificial Intelligence (BIGAI), Beijing, China \\
$^2$ Wangxuan Institute of Computer Technology, Peking University, Beijing, China \\
$^3$ State Key Laboratory of General Artificial Intelligence, Beijing, China \\
\texttt{\{wangyuxuan1,zlzheng\}@bigai.ai} \\ 
}

\begin{document}

\maketitle



\begin{abstract}


Despite progress in multimodal large language models~(MLLMs), the challenge of interpreting long-form videos in response to linguistic queries persists, largely due to the inefficiency in temporal grounding and limited pre-trained context window size.
In this work, we introduce \longname~(\model), a novel framework that bootstraps MLLMs with advanced temporal grounding capabilities and broadens their contextual scope.
Our framework significantly enhances the temporal capabilities of current MLLMs through three key innovations: an efficient multi-span temporal grounding algorithm applied to low-dimension temporal features projected from flow; a multimodal length extrapolation training paradigm that utilizes low-dimension temporal features to extend the training context window size; and a bootstrapping framework that bridges our model with pluggable MLLMs without requiring annotation.
We validate \model across seven video benchmarks and demonstrate substantial performance improvements compared with prior MLLMs. Notably, our model, initially trained on sequences of four frames, effectively handles sequences up to 16$\times$ longer without sacrificing performance, highlighting its scalability and effectiveness in real-world applications. 
Our code is publicly available at \url{https://github.com/bigai-nlco/VideoTGB}.

\end{abstract}



\section{Introduction}
\label{sec:intro}




A fundamental aspect of human intelligence is to effortlessly perceive, memorize, and comprehend daily multi-modal information such as events, observations, and videos that span hours and days.  Such capacity of long-form multi-modal understanding, seamlessly integrating prolonged visual dynamics with textual cues, poses considerable challenges for contemporary machine perceptual systems. A wide range of research works in computer vision and multi-modal tasks has extensively delved into real-life videos, including \ac{videoqa}~\cite{Yu2018AJS,Yu2019ActivityNetQAAD}, text-to-video retrieval~\cite{Hendricks2017LocalizingMI}, video captioning~\cite{Xu2016MSRVTTAL,KrishnaHRLN17}, \etc. Despite the prominent advancements in many video-language benchmarks~\cite{Yu2018AJS,Yu2019ActivityNetQAAD,Hendricks2017LocalizingMI,Xu2016MSRVTTAL,KrishnaHRLN17}, understanding long-form videos with task-oriented linguistic queries still suffers from the significant computational overhead~\cite{Buch2022RevisitingT, Gao2022MISTMI, yu2023selfchained, moviechat, malmm} imposed by high-dimensional video data and the disparity between language and temporal dynamic cues~\cite{Lei2022RevealingSF, nextgqa}.

\begin{figure}[t!]
    \centering
    \includegraphics[width=\linewidth]{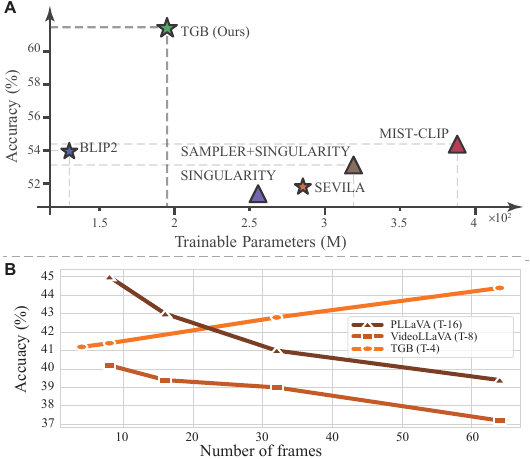}
    \caption{Training Efficiency and Length Extrapolation of \model. \textbf{A.} Results of parameters on AGQA~\cite{GrundeMcLaughlin2021AGQA} Our method demonstrates the best performance with less trainable parameters. \textbf{B.} Results of frame extrapolation on EgoSchema~\cite{egoschema} under zero-shot setting. T-$num$ indicates the number of training context window size. By training with four-frame videos, our model shows consistent performance on extended video length.}
    \label{fig:teaser}
\end{figure}

Some researchers have proposed scaling up the amount of vision data fed into larger models\cite{lvm, lwm}, following the scaling law observed in LLMs. However, the scarcity of high-quality, long video-language datasets makes this approach difficult. Others have explored sampling-based methods to reduce input overhead by selecting relevant frames at either the frame level\cite{Lei2021LessIM,wang2022allinone,Bain2021FrozenIT,Buch2022RevisitingT} or token level~\cite{Gao2022MISTMI}. These methods have three main limitations: first, they are computationally inefficient with slow training and inference speeds due to the large number of tunable parameters; second, the sampling strategy may miss important motion features, especially when there's misalignment between the video segment and the language query; and third, the complexity of the specialized vision encoder complicates the adaptation to long-video understanding.


To address these challenges, we present a novel framework, \longname, which enriches image-language models with temporal priors, significantly improving the understanding of long videos. 
\model distinguishes itself in the following key aspects:

\paragraph{Efficient and Adaptable Video Compression:}
\model features a Bridge that is both lightweight and adaptable. To achieve this, we introduce a learnable multi-span algorithm capable of simultaneously extracting multiple relevant segments from low-dimension motion features. Subsequently, we can compress the entire video into several keyframes. This method efficiently balances performance and resource consumption when processing long-form videos, as demonstrated by our results on the AGQA ~\cite{GrundeMcLaughlin2021AGQA}, with a relatively low parameter count (see ~\cref{fig:teaser}A).

\paragraph{Temporal Extrapolation Preserving Motion Features:}
A significant advantage of the TGB lies in its ability to preserve the continuity of video content, thereby maintaining the temporal dynamics discarded by previously extracted keyframes. To achieve this, we retain the low-dimensional motion features extracted by the TGB to supplement these keyframes. Additionally, we utilize extrapolative position encoding to ensure that these features remain extendable. This approach allows our method to extrapolate to longer sequences in a zero-shot setting (see ~\cref{fig:teaser}B).

\paragraph{Bootstrapping Framework without Annotation:}
Due to the high cost of manual annotations and the limited availability of video data compared to image data, we developed a framework that leverages MLLMs without requiring them to be pretrained on videos. Our approach employs a bootstrapping strategy to refine TGB using MLLMs, eliminating the need for explicit temporal grounding annotations. This strategy also allows for joint training with MLLMs by incorporating the Gumbel-Softmax trick. Additionally, our bootstrapping framework, when integrated with the aforementioned mechanism, can be trained on standard video data and still achieve strong performance on much longer sequences (see ~\cref{fig:teaser}B).

To validate the effectiveness of \model, we conducted experiments on long-form video question answering with seven datasets:  AGQA 2.0~\cite{GrundeMcLaughlin2021AGQA}, NExT-QA~\cite{Xiao2021NExTQANP}, Egoschema~\cite{egoschema},   MSVD~\cite{msvdqa}, MSRVTT~\cite{Xu2016MSRVTTAL}, and ActivityNet~\cite{Yu2019ActivityNetQAAD}. Additionally, we tested temporal question grounding on video using the NExT-GQA dataset~\cite{nextgqa}. Consistent improvements across these datasets confirm the efficacy of our approach. \model has shown strong generalization capabilities across five MLLMs (across encoder, encoder-decoder, and decoder-only) and two LLMs. Further enhancements include the incorporation of a general multimodal instruction-tuning dataset, which shows promise for video chat agent applications. In comparison to other leading-edge methods, \model provides substantial efficiency and efficacy benefits.

\section{Related Work}
\label{sec:related}
\paragraph{Long-form Video Understanding}
The computational demands of processing long-form videos have led to research exploring various methods to address the challenge. A common approach involves sampling-based techniques that aim to reduce the computational load by selectively choosing relevant frames. Research  \cite{Lei2021LessIM, wang2022allinone, Bain2021FrozenIT} integrate sparse sampling within the framework of video-language pretraining. \cite{Buch2022RevisitingT} introduce an atemporal probe (ATP) model that seeks to distill a single image representation from a video clip for more details. Despite these advancements, there's a risk that sparse sampling may lead to an insufficient representation of visual information, which may not be relevant to corresponding language queries. MIST \cite{Gao2022MISTMI} attempts to address this by leveraging the inherent structure of videos to iteratively select and sample spatial-temporal information within a Transformer architecture. Nonetheless, these methods often suffer from reduced computational efficiency and prolonged training and inference times due to the extensive tunable parameters required for processing either spatial or temporal dimensions. More recent studies are exploring the utilization of LLMs for enhancing long-form video understanding. These approaches include a range of techniques such as incorporating temporal embeddings \cite{momentor}, applying prompt-based strategies \cite{yu2023selfchained, timechat}, condensing video frames through a similarity metric \cite{moviechat}, compressing visual tokens with resampling methods \cite{text-resampler, vista-llama, st-llm}, and employing retrieval-based methods that integrate visual features \cite{malmm}. To overcome the constraints of current methods, which usually depend on human-provided annotations for time alignment or require intricate encoding of context, our proposed approach employs a novel bootstrapping framework. This framework enhances a temporal grounding bridge, using MLLMs. This bridge is designed to simultaneously capture multiple granular pieces of key information by leveraging multi-span sampling, which it then integrates with low-dimensional motion features for a more efficient and effective representation.

\paragraph{Bootstrapping Large Language Models for Visual Tasks}
Capitalizing on the success of \acp{llm} in NLP, there is a growing trend of applying them to computer vision tasks, such as VQA~\cite{scienceqa,infoseek,Fu2023MMEAC,liu2023mmbench,Li2023SEEDBenchBM}, image generation~\cite{Ku2023ImagenHubST,Zhang2023MagicBrushAM}, and visual instruction following~\cite{Xu2022MultiInstructIM,Li2023M3ITAL}. The research mainly progresses along three avenues: (i) leveraging \acp{llm}' reasoning for visual tasks~\cite{huang2023language, wu2023visual, driess2023palme, surs2023vipergpt}; (ii) adapting Transformer or linear networks to equip \acp{llm} with visual perception~\cite{li2023blip2, dai2023instructblip, zhu2023minigpt4, xu2023mplug2, gao2023llamaadapter, liu2023visual}; (iii) merging \acp{llm} with video and audio inputs~\cite{zhang2023videollama, maaz2023videochatgpt, lyu2023macawllm}. Recently, Sevila's~\cite{yu2023selfchained} self-chained VideoQA framework uses a two-step approach: selecting keyframes with a tailored prompt and applying them to tasks. However, it faces three issues: time-consuming keyframe localization, static frames missing motion details, and incomplete video representation by sampled frames. Addressing these, we introduce a \model that incorporates both static and dynamic features for video-language understanding.

\begin{figure*}[t!]
    \centering
    \includegraphics[width=\linewidth]{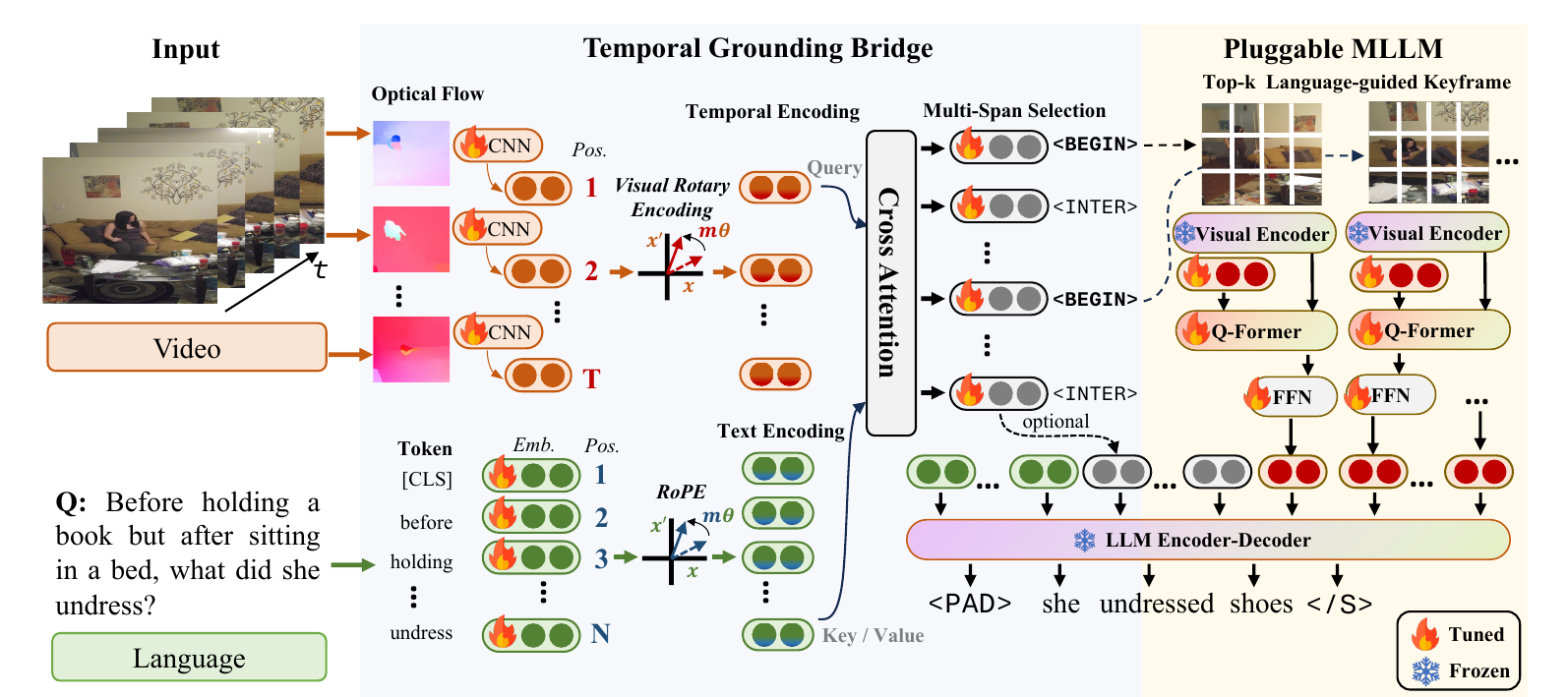}
    \caption{\textbf{Overview of \model framework (BLIP-based).}  The Temporal Grounding Bridge ($\S$\ref{subsec:tps}) is designed to capture temporal priors as well as the specific moments in a video that are grounded by language. We further develop a pluggable bootstraping framework ($\S$\ref{subsec:joint}) that incorporates TGB-MLLM alignment, utilizing a joint optimization strategy. }
    \label{fig:framework}
\end{figure*}

\section{Methodology}
\label{sec:method}



In the subsequent sections, we begin with a detailed formulation of the video-language understanding problem in Section $\S$\ref{sec:problem_def}. Next, in Section $\S$\ref{subsec:tps}, we outline the core components for efficient length extrapolation of our \model. Section $\S$\ref{subsec:joint} explains the process of jointly tuning TGB with pluggable MLLMs on new video-language datasets within our Bootstrapping framework. The overall architecture of \model\ is illustrated in \Cref{fig:framework}.


\subsection{Problem Definition}\label{sec:problem_def}

We formalize the open-ended video-text understanding problem. The input video $V$ is denoted as a sequence of image frames $V=\{fr_1, fr_2, \cdots, fr_{T}\}$, where $T$ is the total number of frames. The input language $L$, denoted as a sequence of $N$ tokens starting with \prompt{[CLS]}, is a task-relevant prompt or question related to interactions among objects, relationships, or events that occur within a few frames of the video. Our goal is to identify the keyframes that relate to the query as grounded moments and generate an open-ended answer in the form of natural language response $y$, incorporating time priors. In the following sections, we use $f^t(\cdot)$ to indicate trainable parameters or neural networks and $f^f(\cdot)$ to indicate frozen pre-trained models.


\subsection{Temporal Grounding Bridge}
\label{subsec:tps}

Previous Video-Language Understanding models commonly extract temporal features from video-text data using offline video encoders or image encoders~\cite{i3dCarreiraZ17, c3dJiangRA17, resnextXieGDTH17, slowfastFeichtenhofer0M19, liu2021Swin, videomaeTongS0022}, causing the model to be time-consuming and lack generality. To address these limitations, we propose a novel mechanism that combines high-dimension key visual cues with low-dimension motion features, ensuring efficiency without compromising visual information.
We further contend that temporal grounding does not necessitate dense frame-level features. To support this claim, we introduce a Temporal Grounding Bridge that incorporates optical flows~\cite{jiang2019stm, slowfastFeichtenhofer0M19,pfister2015flowing,feng2023mutual,zhang2018poseflow} during the temporal grounding stage through a dimensionality reduction. By injecting language queries, this approach generates parameter-efficient, language-guided temporal features. \textbf{A key distinction of our work is that we do not use optical flow merely as supplementary information to enhance frame-based performance. Instead, our framework employs flow as a low-dimensional bridge, which can be directly or indirectly applied to infuse motion details into MLLMs. Importantly, the flow feature can be substituted with other types of features if needed}.

\paragraph{Feature Extraction} 

We denote the optical flow for each pair of video frames as $OF=\{of_1, of_2, \cdots, of_{T}\}$
The low-dimension visual encoding is then computed over these extracted optical flows with a simple convolutional layer followed by a multi-layer perceptron~(MLP) $E_{of} = {\rm MLP^t}({\rm CNN^t}(of))$. For language queries, we use a trainable embedding layer to represent the soft query prompt, \ie, $E_l = Embedding^t(Q)$, where $Q$ is the language query.

\paragraph{Temporal Feature Length Extrapolation}
Despite the impressive efficacy of Transformer-based models within the sphere of deep learning, their operational capacity is inherently constrained by the length of the input. In the context of our research, the bridge is meticulously devised to identify the most salient portions of an entire video, the duration of which may potentially exceed the predetermined limit and differ significantly across various instances. Current literature employs a sampling strategy to condense the video, a process that unfortunately results in the loss of substantial temporal information inherent in the video. To mitigate this challenge, inspired by rotary position embedding (RoPE)~\cite{su2021roformer}, we add multimodal extrapolative position encoding to our TGB(\cref{fig:framework}). Specifically, we compute the position-encoded features using RoPE mechanism for each optical flow and language token, respectively. Formally, the position-encoded features can be denoted as
\begin{align}
    E_{of}^R &= RoPE(W_{of}^tE_{of}, Pos_{of}), \\ E^R_l &= RoPE({W_l}^tE_l, Pos_{l}), \nonumber
\end{align}
where $W_{of}, W_l$ are transformation matrices, $Pos_{of}$, $Pos_{l}$ are corresponding position indices of $OF$ and $L$.

Given the temporal features, we adopt the cross-attention~\cite{vaswani2017attention} mechanism, which is computed using optical flow rotary encoding as query $\mathbf{Q}_R=E_{of}^R$, rotary language embedding as key $\mathbf{K}_R=E_{l}^R$, and language embedding as value $\mathbf{V}=W_VE_l$.

The final language-guided temporal feature $E_R$ is calculated by the standard cross-attention mechanism, \ie, $E_R = {\rm Softmax}(\frac{\mathbf{Q}_R\mathbf{K}_R^{\rm T}}{\sqrt{d_k}})\mathbf{V}$.

\paragraph{Multi-Span Keyframe Selection}
Based on the flow-language encoding, we formulate the temporal question grounding video task as multi-span reading comprehension (RC) problem, where an RC head is to predict the label of fused encoding  $\{{e_R}_1, {e_R}_2, \ldots, {e_R}_T\}$ as one of  \{``\prompt{<BEGIN>}'', ``\prompt{<END>}'',  ``\prompt{<NONE>}''\} of the grounded video spans. The selection can be formulated as:
\begin{align}
    h & = \mathcal{F}^t_{\theta}({e_R}_1, {e_R}_2, \ldots, {e_R}_T), \\ \nonumber
    index & = \argmax ({\rm Softmax}(h)),
\end{align}
where $\mathcal{F}_\theta$  denotes the RC head for span selection,  $index$ is the prediction of the start or end index. The objective is computed as the cross-entropy between the prediction and pseudo labels. 
During Inference, we can obtain an arbitrary number of $k$ segments of grounded video by predicting $k$ \prompt{<BEGIN>}s and $k$ \prompt{<END>}s with the RC Head. Finally, we union these segments to eliminate the overlap between these extracted spans.  


\begin{figure}[H]
    \centering
    \includegraphics[width=\linewidth]{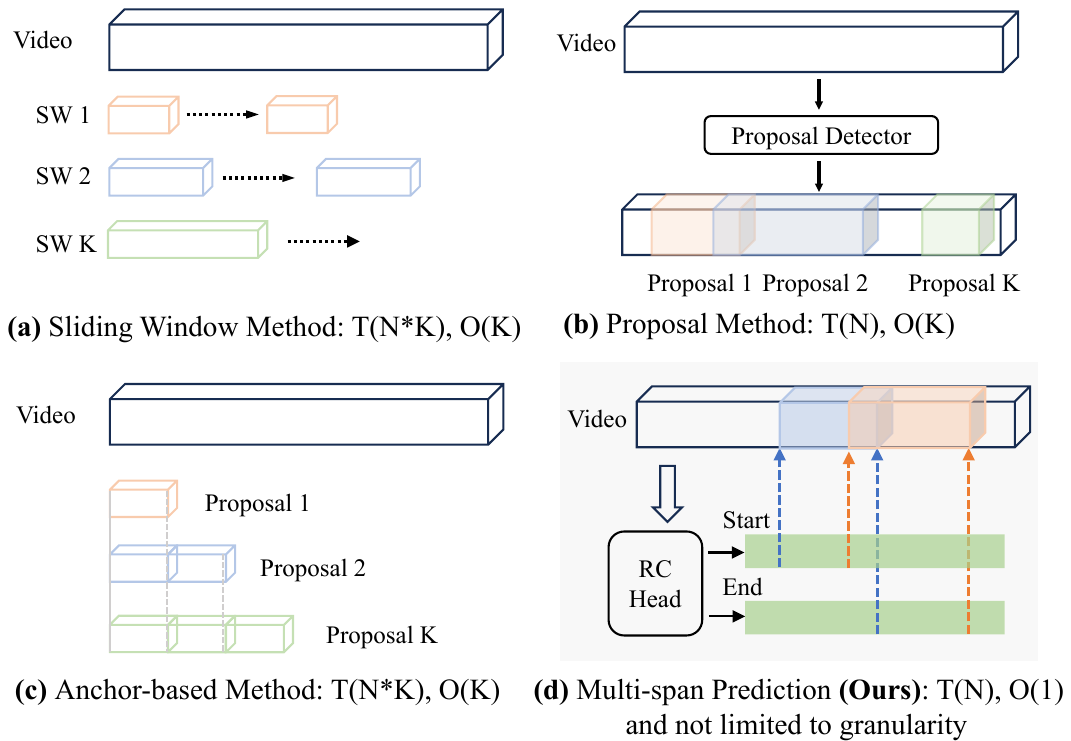}
    \caption{\textbf{Comparison of multi-span RC prediction (d)} and other methods (a-c) in terms of time and space complexity.}
    \label{fig:multispan}
\end{figure}
In ~\cref{fig:multispan}, we compare our proposed multi-span reading comprehension prediction algorithm and other commonly used methods for temporal sentence grounding on video tasks, including the sliding window method, proposal method, and anchor-based method. Compared with other span-fixed methods, our method could obtain multiple grounded video spans with the least time complexity and space complexity.

\paragraph{Bridge with MLLMs}

For each selected keyframe $fr_k$, we utilize a frozen pre-trained visual encoder to capture its spatial information, \ie, $E_{fr} = Enc_v^f(fr_k)$.
In line with contemporary research, we adapt the visual feature via a pre-trained Q-former and obtain $q$ query representations. $\Tilde{E_q} = Enc_q^t(E_q^t, E_{fr})$
, where $E_q^t$ represents the learnable query, $\Tilde{E_q} = \{{e_q}\}$ is the spatial visual feature output of the MLLM. The final output is produced by feeding obtained spatial-temporal-language information in to a forzen \ac{llm}, \ie, $y={\rm LLM^f}(E_r, \Tilde{E_q}, E_l)$.

\subsection{Joint Training Bootstrapping Framework}
\label{subsec:joint}

\paragraph{Bootstraping Algorithm}
Due to the scarcity of video-language datasets with temporally grounded annotations and the high cost of acquiring human labeling, we have developed a self-improvement algorithm to enhance TGB using the capabilities of MLLM. There are two primary types of video-language understanding tasks: close-ended and open-ended. We have tailored algorithms to address both types.
For close-ended tasks, we employ an iterative method in which each video frame is evaluated using the MLLM. Frames that lead to correct MLLM predictions are marked with positive labels, while those with incorrect predictions receive negative labels.
For open-ended tasks, which often lack temporal labels, we introduce an innovative approach to generate pseudo labels for open-ended datasets. We analyze the MLLM-generated results of uniformly sampled frames and compute the sentence similarity between these results and the ground truth. We then apply a monotonic stack algorithm to identify the span with the highest similarity scores. These pseudo labels are used to optimize the TGB. Detailed information about the this algorithm can be found in the \textbf{Appendix \ref{supp:pseudo-label}}.

\paragraph{Joint Optimization}
Despite the utilization of pseudo labels in the training process, in many videos, there is implicit alignment between query and videos. In addition, the fixation of the pre-trained bridge within the bootstrapping framework inevitably leads to the introduction of exposure bias. To mitigate this we suggest a joint training approach that extends the Gumbel-Softmax technique. We implement Gumbel-Softmax sampling $K$ times to sample $K$ spans:
\begin{equation}
    {\rm GumbelSoftmax}(\mathcal{F}_{\theta}^t({e_R}_1,\ldots,{e_R}_T), \tau),
\end{equation}
where $\tau$ is the scaling term for reparameterizing. Consequently, our methodology is employed to facilitate a connection between TGB and MLLMs, thereby enabling our framework to be jointly optimized on domain-specific datasets.

\section{Experiments}
\label{sec:exp}

In this section, we utilize the \model on 5 MLLMs, across encoder, encoder-decoder, and decoder-only three types of architectures. We demonstrate the effectiveness of our approach on three tasks: long-form videoQA and zero-shot open-domain videoQA (\cref{sec:videoqa}), temporal question grounding on video (\cref{exp:nextgqa}). Furthermore, We provide a detailed analysis to showcase the effectiveness of our framework in length extrapolation (\cref{fig:teaser}B), the effectivness of different components (\cref{sec:ablate}), and compare its computational efficiency with other state-of-the-art models on a similar scale (\cref{sec:time}).


\subsection{Long-form Video Question Answering}
\label{sec:videoqa}

\begin{table*}[ht!]
    \centering
    
    \resizebox{\linewidth}{!}{%
    \begin{tabular}{l|cccccccc|c}
    \toprule
        \textbf{Model} & \textbf{\makecell{Object-\\relation}} & \textbf{\makecell{Relation-\\action}} & \textbf{\makecell{Object-\\action}} &  \textbf{Superlative} &  \textbf{Sequencing} & \textbf{Exists} & \textbf{\makecell{Duration \\comparison}} & \textbf{\makecell{Action\\recognition}} & \textbf{Overall} \\ \midrule
        \multicolumn{10}{l}{\textit{Retrieval-based Video-Language Models}} \\

        HME~\cite{Fan2019HeterogeneousME} & 37.42 & 49.90 & 49.97 & 33.21 & 49.77 & 49.96 & 47.03 & 5.43 & 39.89 \\
        
        PSAC~\cite{DBLP:conf/aaai/LiSGLH0G19} & 37.84 & 49.95 & 50.00 & 33.20 & 49.78 & 49.94 & 45.21 & 4.14 & 40.18 \\ 

        HCRN~\cite{DBLP:conf/cvpr/LeLV020} & 40.33 & 49.86 & 49.85 & 33.55 & 49.70 & 50.01 & 43.84 & 5.52 & 42.11 \\ 

        AIO~\cite{wang2022allinone} & 48.34 & 48.99 & 49.66 & 37.53 & 49.61 & 50.81 & 45.36 & 18.97 & 48.59 \\
        ATP~\cite{Buch2022RevisitingT} & 50.15 & 49.76 & 46.25 & 39.78 & 48.25 & 51.79 & 49.59 & 18.96 & 49.79 \\
        
        MIST-AIO~\cite{Gao2022MISTMI} & 51.43 & 54.67 & 55.37 & 41.34 & 53.14 & 53.49 & 47.48 & 20.18 & 50.96 \\ \midrule
        


        ALBEF & 50.53 & 49.39 & 49.97 & 38.22 & 49.79 & 54.11 & 48.01 & 10.40 & 50.68 \\

        \textbf{ALBEF + TGB (Ours)} & 51.05 & 51.11 & 51.66 & 38.36 & 51.33 & 58.10 & 49.20 & 11.78 & 51.73 \\

        SINGULARITY~\cite{Lei2022RevealingSF} & 50.87 & 50.67 & 49.70 & 40.47 & 40.79 & 55.34 & 48.20 & 11.59 &  51.11 \\

        \textbf{SINGULARITY + TGB (Ours)} & 52.33 & 54.12 & 55.07 & 40.71 & 54.49 & 57.88 & 48.35 & 12.24 & 53.13 \\  

        VIOLET~\cite{Fu2021VIOLETE} & 50.89 & 50.24 & 50.93 & 40.76 & 50.51 & 58.07 & 38.97 & 6.53 & 51.03 \\
        
        \textbf{VIOLET + TGB (Ours)} & 51.59 & 54.54 & 56.96 & 40.94 & 55.61 & 59.12 & 42.81 & 9.02 & 52.59 \\

        \midrule

        \multicolumn{10}{l}{\textit{Open-ended Video-Language Models}} \\

        SeViLA$^*$~\cite{yu2023selfchained} & 51.15 & 48.93 & 62.08 & 42.24 & 55.96 & 53.02 & 38.91 & 0.00 & 51.70 \\
        
        BLIP2~\cite{li2023blip2}  & 53.72 & 48.64 & 62.1 & 43.84 & 55.94 & 55.14 & \textbf{40.39} & \textbf{0.28} & 54.00 \\ 

        \textbf{\model-BLIP2~(Ours)} & \textbf{62.27} & \textbf{51.74} & \textbf{66.09} & \textbf{53.67} & \textbf{60.11} & \textbf{60.85} & 36.99 & 0.00 & \textbf{61.45} \\

         \bottomrule
         \multicolumn{10}{p{\dimexpr\textwidth\relax}}{\footnotesize $^*$ Re-implementation result. We removed prior information from QVHighlights~\cite{leidetecting} used in SeViLA for fair comparison.}
    \end{tabular}
    }
    \caption{\textbf{Comparison accuracy of different sampling-based SOTA models on AGQA 2.0}.}
    \label{tab:res_blipagqa2}
\end{table*}

\begin{table}[ht!]
\centering
    \resizebox{\linewidth}{!}{%
    \begin{tabular}{l|ccc|c}
    \toprule

        \textbf{Model} & \textbf{Temporal} & \textbf{Causal} & \textbf{Description} & \textbf{All} \\ \midrule

        \multicolumn{4}{l}{\textit{Retrieval-based Video-Language Models}} \\

        CLIP~\cite{pmlr-v139-radford21a} & 46.3 & 39.0 & 53.1 & 43.7 \\
        
        HGA~\cite{DBLP:conf/aaai/JiangH20} & 44.2 & 52.5 & 44.1 & 49.7 \\

        AIO~\cite{wang2022allinone} & 48.0 & 48.6 & 63.2 & 50.6 \\ 

        VQA-T~\cite{yang2021justask} & 49.6 & 51.5 & 63.2 & 52.3 \\

        MIST-AIO~\cite{Gao2022MISTMI} & 51.6 & 51.5 & 64.2 & 53.5 \\

        ATP~\cite{Buch2022RevisitingT} & 50.2 & 53.1 & 66.8 & 54.3 \\ 

        VGT~\cite{DBLP:conf/eccv/XiaoZCY22} & 52.3 & 55.1 & 64.1 & 55.0 \\
        
        MIST-CLIP~\cite{Gao2022MISTMI} & 56.6 & 54.6 & 66.9 & 57.1 \\\midrule

        \multicolumn{4}{l}{\textit{Open-ended Video-Language Models}} \\

        BLIP2~\cite{li2023blip2} & 64.9 & 69.7 & 79.4 & 69.6 \\

        SeViLA$^*$~\cite{yu2023selfchained} & 66.4 & 71.9 & 80.8 & 71.5 \\
        

        \textbf{\model-BLIP2 (Ours)} & \textbf{66.5} & \textbf{72.8} & \textbf{81.2} & \textbf{72.1} \\

         \bottomrule
         \multicolumn{5}{c}{\makebox[\linewidth][c]{\footnotesize $^*$ We removed prior information from QVHighlights used in SeViLA for fair comparison.}}

    \end{tabular}
    }
    \caption{\textbf{Comparison accuracy of long-form video QA on NExT-QA}.}\label{tab:res_blipnextqa}
\end{table}

\paragraph{Setups}
We take three long-form \ac{videoqa} benchmarks AGQA~\cite{GrundeMcLaughlin2021AGQA}, NExTQA~\cite{Xiao2021NExTQANP}, and EgoSchema~\cite{egoschema} for evaluation.  We use two types of baselines: retrieval-based models and open-ended models focusing on recent SOTA temporal priors learning models for comparative analysis. For the retrieval-based models, in addition to traditional methods~\cite{Fan2019HeterogeneousME, DBLP:conf/aaai/LiSGLH0G19,DBLP:conf/cvpr/LeLV020,wang2022allinone,Li2021AlignBF,Lei2022RevealingSF,Fu2021VIOLETE}, we use recent SOTA temporal learning models, specifically ATP~\cite{Buch2022RevisitingT} and MIST~\cite{Gao2022MISTMI}.
For the open-ended models, we use BLIP2~\cite{li2023blip2} and SEVILA~\cite{yu2023selfchained}. For the number of keyframes, we sample 4 frames for \model and 6 frames for TGB-augmented methods (where we don't incorporate the motion feature to the input directly) in all experiments. For more implementation details, please refer to \textbf{Appendix ~\ref{supp:implementation}}.

\paragraph{Results on AGQA 2.0}

Our \model framework, compared with prior works that integrate keyframe localization into video-language tasks, shows that BLIP2, despite its 4.1B parameters pre-trained on 129M images, offers only a slight improvement over smaller models, as demonstrated in AGQA 2.0 results. BLIP2 even falls short of the state-of-the-art MIST-CLIP, which has a parameter count comparable to BERT \cite{Devlin2019BERTPO}. This indicates that simply adapting videos for \acp{llm} is inadequate for complex video question-answering tasks. However, when enhanced with our \model framework, BLIP2's accuracy increases by $7.45$ points, underscoring the framework's ability to learn spatial-temporal video features effectively. We believe this is due to our framework's superior temporal information capture, which other methods miss. Nonetheless, it still lags behind MIST-CLIP on certain question types, stemming from the inherent differences in how retrieval-based and open-ended models produce answers. For example, open-ended models struggle with ``Duration comparison" questions because they are limited to generating answers from a specific set of 171 words or phrases, which are infrequently found in generative models' pre-training data, posing a challenge for exact match generation.


\begin{table}[t!]
    \centering
    
    \resizebox{\linewidth}{!}{
    \begin{tabular}{l|c|c|c}
        \toprule
        \textbf{Methods} & \textbf{Base Model} & \textbf{\# of Frames} & \textbf{Accuracy} \\\midrule
        Sevila & BLIP2 & 32 & 25.7  \\ 
        mPLUG-Owl & LLaMA-7b & 5 & 33.8 \\
        Video-LLaVA & LLaVA-7b & 8 & 40.2 \\
         
         \textbf{\model-BLIP2} & BLIP2& 4 & \textbf{41.2} \\
         \bottomrule
    \end{tabular}

    }
    \caption{\textbf{Zero-shot Result on subset of EgoSchema}}
    \label{tab:egoschema}
\end{table}

\begin{table}[t!]
    \centering
    
    \resizebox{\linewidth}{!}{
    \begin{tabular}[0.4\textwidth]{l|c|cc|cc|cc}
        \toprule
         \multirow{2}{*}{\textbf{Methods}} & \multirow{2}{*}{\textbf{LLM size}} & \multicolumn{2}{c|}{\textbf{MSVD-QA}} & \multicolumn{2}{c|}{\textbf{MSRVTT-QA}} & \multicolumn{2}{c}{\textbf{ActivityNet-QA}}   \\ 
         & & Accuracy & Score & Accuracy & Score & Accuracy & Score \\ \midrule
         FrozenBiLM & 1B & 32.2 & - & 16.8 & -  & 24.7 & - \\
         VideoChat & 7B & 56.3 & 2.8 & 45.0 & 2.5  & - & 2.2\\
         LLaMA-Adapter & 7B & 54.9 & 3.1 & 43.8 & 2.7 & 34.2 & 2.7 \\
         Video-LLaMA & 7B & 51.6 & 2.5 & 29.6 & 1.8  & 12.4 & 1.1 \\
         Video-ChatGPT & 7B & 64.9 & 3.3 & 49.3 & 2.8  & 35.2 & 2.7 \\
        \textbf{\model (BLIP2)}  & 3B & 66.0 & 3.6 & 53.5 & 3.1 & 41.3 & 3.1 \\
         \textbf{\model (Vicuna7B)} & 7B & \textbf{71.4} & \textbf{3.9} & \textbf{57.3} & \textbf{3.3} & \textbf{43.9} & \textbf{3.3} \\                
         \bottomrule
    \end{tabular}
    }
    \caption{\textbf{Zero-shot Open Domain Video QA.}}
    \label{tab:opendomain}
\end{table}

\paragraph{Results on NExTQA} 
\Cref{tab:res_blipnextqa} presents the results on the NExTQA dataset.
Generally, our method outperforms a variety of baselines, particularly SeViLA, a recent model using LLM for keyframe selection. However, the performance improvement of our framework on NExTQA is not as significant as on AGQA. This is because NExTQA places more emphasis on causality, and videos in NExTQA, sourced from VidOR \cite{shang2019annotating, thomee2016yfcc100m}, a dataset focused on video objects and relation recognition, exhibit more "static appearance bias"~\cite{Lei2022RevealingSF} than AGQA.

\paragraph{Results on EgoSchema} We evaluated our model's performance on the EgoSchema \cite{egoschema}, one of the longest videoQA datasets available. We apply this experiment under the zero-shot setting, thereby trained on video instruction dataset from VideoLLaVA~\cite{Lin2023VideoLLaVALU}. As shown in \cref{tab:egoschema}, our model outperforms other models that use similar pretraining data. This superior performance is particularly notable given that our base model is smaller and processes fewer input instances compared to the others. We believe our approach is highly effective for understanding long-form video content.

\paragraph{Impact of TGB-grounded frames} We assessed the influence of TGB on different MLLMs by testing them with alternative MLLMs and TGB-grounded frames, excluding optical flow features. For MLLMs using single-image input, we merged multiple images using an early fusion approach. Our experiments on the AGQA 2.0 dataset in \cref{tab:res_blipagqa2} revealed: \one \textit{TGB matters in temporal learning over different MLLMs.} TGB-augmented methods significantly enhances MLLMs' ability in solving temporal question  (\ie, ``\prompt{Relation-action}", ``\prompt{Sequencing}", ``\prompt{Exists}") compared to the uniform sampling strategy. \two \textit{Absence in temporal priors hinders the performance of ensemble methods.} The improvement gained on SINGULARITY is better than ALBEF, despite they have similar objectives but SINGULARITY is pre-trained with video corpora. \three \textit{Temporal features of optical flow can compensate for the information loss caused by frame sampling.}  The marginal improvement of our TGB-augmented models on ``\prompt{Superlative}" suggests that the sampling strategy cannot enhance the model's overall video understanding ability. In contrast, our BLIP2-based framework with optical flow improves from 43.84 to 53.67 (a relative increase of $22.42\%$).
indicating that optical flow features can reduce the temporal information loss caused by the sampling strategy.

\paragraph{Analysis of Pluggable MLLMs}
We substitute the BLIP2 with three popular types of MLLMs, mainly encoder-based models, \ie, VIOLET~\cite{Fu2021VIOLETE} as a representative of video-language models, ALBEF~\cite{Li2021AlignBF} as an image-language model, SINGULARITY~\cite{Lei2022RevealingSF} as a pre-trained model on a single frame of video and image corpus. It's noteworthy that we did not incorporate the learned optical flow feature into these MLLMs' input. In this part, we also apply all the experiments on AGQA 2.0 dataset. \Cref{tab:res_blipagqa2} (ALBEF + TGB, VIOLET + TGB, SIGULARITY + TGB) validates the efficacy of our TGB and the versatility of our framework. On average, the solver achieves a $3.68\%$  accuracy improvement after replacing the uniform sampled frames with keyframes extracted by the TGB. These results consistently demonstrate the effectiveness of our \model framework across various MLLMs.

\paragraph{Generality of \model} To demonstrate the generality of our approach, we applied our model to visual instruction datasets~\cite{Lin2023VideoLLaVALU}. We also adapted the LLM using LoRA~\cite{DBLP:conf/iclr/HuSWALWWC22} to ensure a fair comparison with current SOTA methods. As shown in Table~\ref{tab:opendomain}, our method's performance on the videoQA dataset in a zero-shot setting is presented. Unlike VideoLLaVA, our method was not pretrained on additional datasets; it was only fine-tuned on the same visual instruction datasets. The results demonstrate that our method can match the performance of the latest state-of-the-art (SOTA) MLLMs, even though the LLM of our model is less than half their size. This highlights the considerable promise of our framework in this domain.

\subsection{Temporal Question Grounding on Video}
\label{exp:nextgqa}

\begin{table}[t!]
    \centering
    
    \resizebox{\linewidth}{!}{
    \begin{tabular}[\linewidth]{l|c|ccc}
        \toprule
         \textbf{Method} & \textbf{Vision Encoder} & \textbf{mIoU} & \textbf{IoU@0.3} & \textbf{IoU@0.5}    \\ \midrule
         VGT & RCNN & 3.0 & 4.2 & 1.4 \\
         VIOLETv2 & VSWT & 3.1 & 4.3 & 1.3  \\
         Temp[Swin] & SWT & 4.9 & 6.6 & 2.3  \\
         Temp[CLIP] & ViT-B & 6.1 & 8.3 & 3.7  \\
         Temp[BLIP] & ViT-B & 6.9 & 10.0 & 4.5  \\
         FrozenBiLM & ViT-L & 7.1 & 10.0 & 4.4  \\ 
         IGV & ResNet & 14.0 & 19.8 & 9.6  \\
         \textbf{\model} & OF+CNN & \textbf{19.9} & \textbf{23.3} & \textbf{11.2} \\
         \bottomrule

         
    \end{tabular}
    }
    \caption{\textbf{Comparison results of Temporal Question Grounding task on NExT-GQA~\cite{Xiao2023CanIT}.} }
    \label{tab:grounding}
\end{table}

\paragraph{Setup}
We use the Temporal Question Grounding on Video (TQGV) dataset NExT-GQA~\cite{nextgqa} to evaluate the efficacy of our TGB. We select a wide range of MLLMs as baselines: VGT~\cite{DBLP:conf/eccv/XiaoZCY22}, Temp~\cite{Buch2022RevisitingT,Xiao2023CanIT}, FrozenBiLM~\cite{DBLP:conf/nips/YangMSLS22}, IGV~\cite{DBLP:conf/cvpr/0004WXJC22}, and SeViLA~\cite{yu2023selfchained}. These baseline models encompass a variety of architectures, text encoders, and vision encoders. In contrast, our method does not depend on heavy offline vision feature extractors. We obtain the optical flow using a fixed RAFT \cite{2020raft}, a model with only 5.26 million parameters. This comparison highlights the efficiency and simplicity of our approach.

\paragraph{Main Results and Analysis}

As shown in~\cref{tab:grounding}, our method outperforms baselines using additional feature extractors~\cite{Ren2015FasterRT,Liu2021VideoST,liu2021Swin,pmlr-v139-radford21a}. Our TGB with optical flow effectively learns temporal priors for video-language tasks. We suggest that discrete frames may introduce irrelevant visual cues, increasing the computational load for temporal learning. Despite this, all methods struggle with temporal grounding, with most mIoU values under $0.20$, indicating a significant gap in current temporal modeling. Conversely, our TGB's temporal features could mitigate these issues. We posit that our approach could significantly advance spatial-temporal research for extended videos. Qualitative results are presented in Appendix ~\ref{supp:qualitative-nextgqa}.

\subsection{Ablation Study}
\label{sec:ablate}
    




        


\begin{table}[t!]
    \centering
    
    \resizebox{\linewidth}{!}{%
    \begin{tabular}{l|cccc|c}
    \toprule
        \textbf{Model} & \textbf{\makecell{Object-\\relation}} & \textbf{\makecell{Relation-\\action}} & \textbf{\makecell{Object-\\action}} &  \textbf{Others}  & \textbf{All} \\ \midrule

        \model & 62.27 & 51.74 & 66.09 & 57.04 & 61.45 \\

        ~~w/o optical flow & 59.13 & 15.06 & 50.79 & 51.29 & 55.00 \\
        
        ~~w/ fixed bridge & 62.28 & 47.84 & 50.68 & 53.47 & 59.88 \\

        ~~w/ uniform sampling & 53.72 & 48.64 & 62.10 & 50.68 & 54.00 \\

        ~~w/ zero-shot & 23.60 & 17.09 & 29.37 & 40.72 & 25.54 \\
                    
         \bottomrule
    \end{tabular}
    }
    \caption{\textbf{Ablation study of our method on reasoning questions from AGQA 2.0.} We list the major outputs of complicated relationships and summarize the rest; see \textit{SM} for complete results. }
    \label{tab:ab_blipagqa2}
    \vspace{-.1in}
\end{table}

We apply ablation study on \model to investigate the effects of our joint training framework. All the experiments are performed on AGQA 2.0~\cite{GrundeMcLaughlin2021AGQA}. As shown in \Cref{tab:ab_blipagqa2}, the framework incorporating motion feature significantly improved performance by $11.72\%$, underscoring its effectiveness in tackling spatial-temporal problems. We also found that fixing the pre-trained TGB during training notably affected performance on temporal questions like ``\prompt{Relation-action}", suggesting that joint training can further optimize the bridge. Lastly, comparing with zero-shot and fine-tuned BLIP2~\cite{li2023blip2} with uniformly-sampled frames, our method showes significant improvements, demonstrating its overall effectiveness. In~\cref{supp:tsp-ablation}, we provide detailed ablation study about the TGB-augmented models.


\begin{table}[h]
    \centering
    
    \begin{tabular}[\linewidth]{l|ccc}
        \toprule
         \textbf{Method} &   \textbf{mIoU}    \\ \midrule
         Sliding Window  & 17.65  \\
         Proposal  & 14.09   \\
         Anchor-based & 14.20  \\
         Multi-span Prediction (Ours) & 19.9   \\
         \bottomrule
    \end{tabular}
    \caption{\textbf{Comparison of multi-span prediction and other methods on NExT-GQA dataset}}
    \label{tab:mc-groudning}
\end{table}

In \cref{fig:multispan}, we demonstrate the superiority and efficiency of our method. In this section, we will reveal the efficacy of our proposed multi-span method compared to other methods for the temporal sentence grounding task in video. We conduct additional experiments on the NExT-GQA~\cite{nextgqa} dataset, comparing different grounding strategies using the mIoU metric. The results are shown in \cref{tab:mc-groudning}, where our methods exhibit a significant performance improvement over other methods.

\subsection{Time Efficiency}
\label{sec:time}

\begin{table}[h]
    \centering
    
    \resizebox{\linewidth}{!}{
    \begin{tabular}[\linewidth]{l|ccc}
        \toprule
         \textbf{Model} &  \makecell{\textbf{FLOPs} \\ (GFLOPs)} $\downarrow$ & \makecell{\textbf{MACs} \\ (GMACs)} $\downarrow$ & \textbf{Acc.} $\uparrow$   \\ \midrule
         BLIP2 (ViT-G)  & 2,705 &  1,350 & 69.6  \\
         Sevila (ViT-G)  & 13,720 & 14,357 & 71.5   \\
         \model(ViT-G) & 19,620 & 9,840 & 72.3  \\
         \model(OFs) & 2,950& 1,474 & 72.1   \\
         \bottomrule
    \end{tabular}
    }
    \caption{\textbf{Computational Efficiency of \model. }}
    \label{tab:efficiency}
    \vspace{-.15in}
\end{table}

We evaluated the average inference time efficiency of our method against BLIP2 using calflops~\cite{calflops} on the NExT-QA dataset, as shown in \Cref{tab:efficiency}. Our method outperformed the current SOTA model SeViLa, which uses the LLM to select keyframes, both in terms of performance and efficiency. While replacing the OFs with features from ViT-G~\cite{Zhai2021ScalingVT} resulted in minor improvements, it significantly increased computation costs due to the offline feature extractor. Compared to BLIP2, our method required minimal additional computation. The major computation costs were associated with the \acp{llm} from BLIP2 and the offline feature extractor. We believe our method strikes a balance between being effective and computationally efficient. Further details on the composition of the inference time of \model are provided in \textit{SM}. 



\begin{figure}[H]
    \centering
    \includegraphics[width=\linewidth]{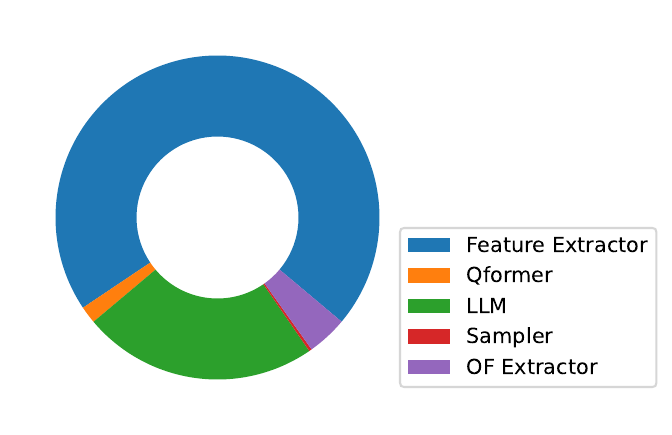}
    \caption{Inference time Analysis}
    \label{fig:infer_time}
    \vspace{-.15in}
\end{figure}

We further investigate the composition of inference time of \model on the NExT-QA dataset. We find most computation costs come from LLM and the offline feature extractor. Compared with other components, the computation cost is trivial, indicating the strong efficiency of our method. The offline demo is presented in the supplementary material.

\section{Qualitative Studies on NExTGQA}
\label{supp:qualitative-nextgqa}
\begin{figure}[H]
    \centering
    \includegraphics[width=\linewidth]{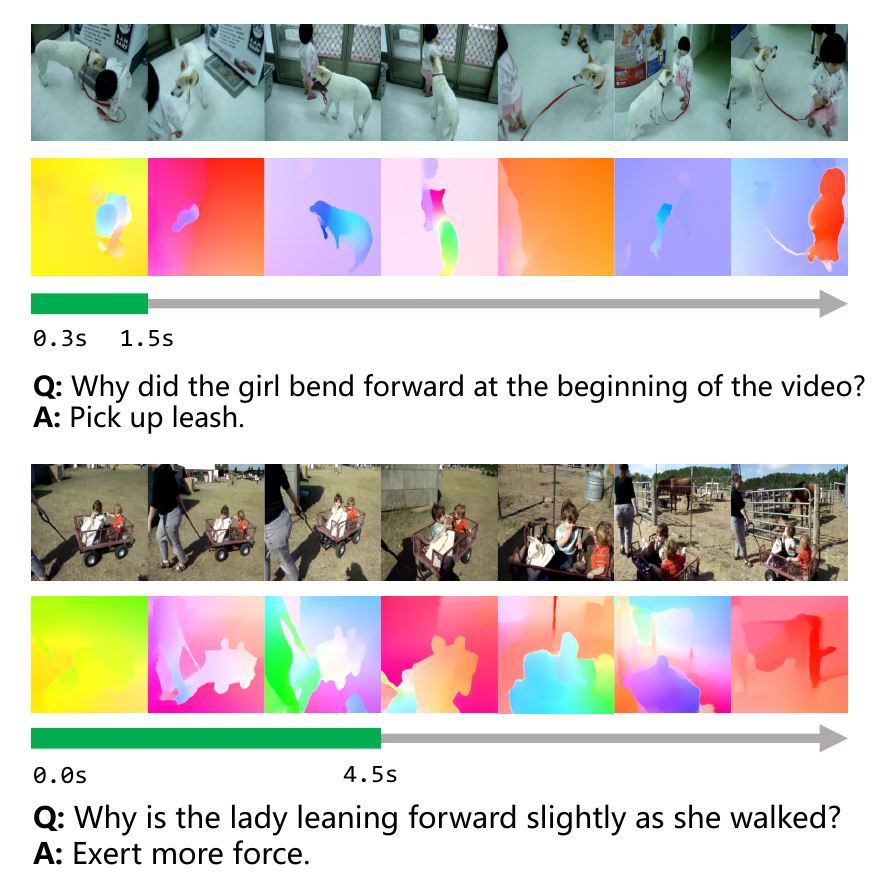}
    \caption{Qualitative results on temporal grounding}
    \label{fig:grounding}
    \vspace{-.1in}
\end{figure}

\cref{fig:grounding} presents two random outputs from \model on the TQGV task. The first example demonstrates how our method can ground video using the semantic information from the question, specifically, the phrase ``\prompt{at the beginning}". The second example demonstrates the efficacy of our method in temporal reasoning, as evidenced by the phrase ``\prompt{as she walked}".

\section{Conclusion}
\label{sec:conclusion}

In this work, we propose a pluggable framework \model for long Video-Language Understanding tasks, which comprises a TGB and a spatial prompt solver to combine spatial-temporal-language alignment and temporal grounding. 
Experiments on long-form video question answering and temporal question grounding on video demonstrate a consistent improvement over various types of MLLMs. Comprehensive analysis verifies the effectiveness, efficiency, and generality of our framework. 


\section*{Limitations}

Our study has one primary limitation:
\ie \textbf{Limited Temporal Grounding Capability} As shown in \cref{exp:nextgqa}, our method outperforms existing approaches but still has restricted temporal grounding capabilities, a common issue in current research. We suspect that this limitation may be due to the constraints of the lightweight 6-layer transformer-based TGB. In future work, we aim to enhance this aspect of our method without sacrificing efficiency.

\clearpage


\section*{Acknowledgements}

The authors thank the reviewers for their insightful suggestions to improve the manuscript. This work presented herein is supported by the National Natural Science Foundation of China (62376031). 

\bibliography{custom}

\clearpage
\appendix

{\bf\Huge Appendices \bigskip}


    

\DoToC

\section{Self-Improvement Algorithm}
\label{supp:pseudo-label}
\cref{alg:label} shows our self-improvement algorithm of automatically generating pseudo labels by the MLLM, which is used to optimize the TGB.

\begin{algorithm}[h!]
  \SetAlgoLined
  \DontPrintSemicolon
  \SetKwComment{Comment}{/* }{ */}
\caption{Pseudo Label Algorithm}
\label{alg:label}
\KwIn{frames ($V=\{fr_1, fr_2, \cdots, fr_{T}\}$), query ($q$), answer ($a$)}
\KwOut{temporal grounded span}
\SetKwFunction{ExoViP}{ExoViP}
$score_{best}$ $\gets$ $0$\;
$start$ $\gets$ $0$\;
$end$ $\gets$ $T-1$\;
$stack$ $\gets$ $empty\ list$\;
$scores$ $\gets$ $empty\ list$\;
\For{$fr$ in $V$}{
    $prediction = LLM_{MLLM}(fr, q)$\;
    $scores.add(SIM(prediction, a))$
}
\For{$i$ in $scores.length$}{
    \While{$stack$ is not empty and $stack.get(score.top)>score.get(i)$ }{
        $tmp = stack.pop()$\;
        $score_{tmp} = (i - stack.top -1 ) \times score.get(tmp)$\;
        \eIf{$score_{tmp} > score_{best}$}{
            $score_{best} = score_{tmp}$\;
            $start=0$\;
            $end=i-2$
        }{}
    }
    $stack.push(i)$
}
\end{algorithm}




\section{More Analysis Experiments}
\label{sec:ablation-analysis}

\subsection{Ablated TSP-augmented models} 
\label{supp:tsp-ablation}

\begin{table}[H]
    \centering
    
    \resizebox{\linewidth}{!}{
    \begin{tabular}[0.4\textwidth]{l|c|c|c|c}
        \toprule
         \textbf{TGB} & \textbf{MLLM} & \textbf{\makecell{\# of frames \\ (Train)}} & \textbf{\makecell{\# of frames \\ (Infer.)}} & \textbf{Acc.}   \\ \midrule
         OF & SING-17M & 1 & 6 & 53.13 \\
         OF & SING-17M & 1 & 1 & 51.36 \\
         OF & SING-17M & 6 & 6 & 53.85 \\
         OF & SING-5M & 1 & 6 & 51.10 \\
         Swin. & SING-17M & 1 & 6 & 53.76 \\
         \bottomrule
    \end{tabular}
    }
    \caption{Detailed Analysis on the TGB.}
    \label{tab:ablation}
\end{table}

In ~\cref{tab:ablation}, we analyzed TSP+SINGULARITY to evaluate the TSP-augmented paradigm. Our study revealed that increasing the number of frames during inference improved performance by $3.4\%$, but further increases did not proportionally enhance results. We also found that MLLM benefits more from the sampling strategy when adequately pretrained (\ie, 17M denotes the model is pretrained on 17M video corpora). Additionally, we proposed two TGB variants, replacing optical flow with features extracted by the video SwinTransformer~\cite{Liu2021VideoST} for pre-training. The comparable results suggest that our TSP can effectively reason over time without any prior perception information.

\subsection{Influence of the number of frames on solver}
\label{analysis}

\begin{figure}[H]
    \centering
    \includegraphics[width=.8\linewidth]{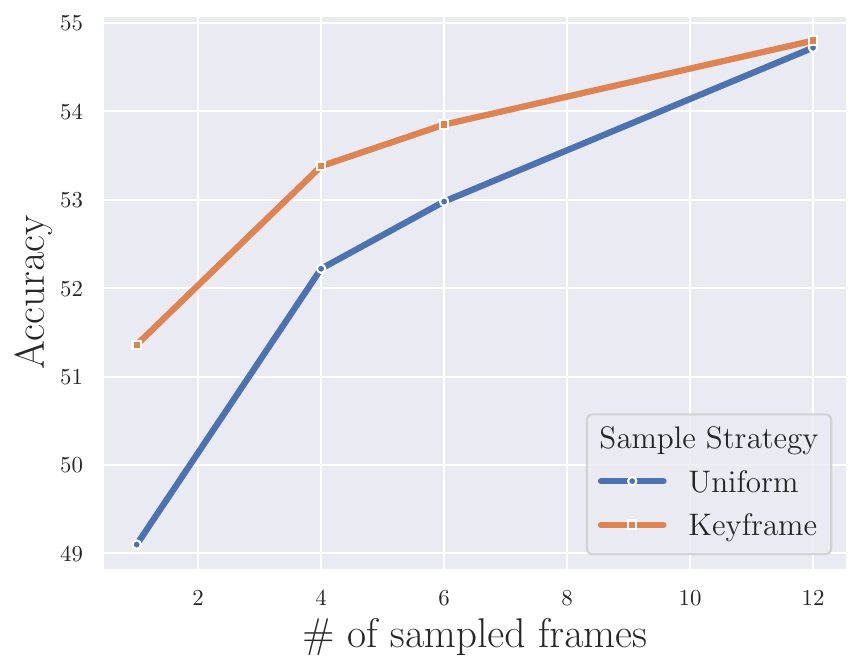}
    \caption{Further study on the number of sampled frames.}
    \label{fig:frames}
\end{figure}

We trained the solver with different numbers of sampled frames. Results are shown in \Cref{fig:frames}. The fewer sampled frames the better performance of the keyframe strategy, and after a certain point, the uniform strategy performs close to the keyframe strategy. This is because the average duration of videos in AGQA is around 30 seconds, 12 frames are close to dense sampling which covers almost all visual cues. In other words, video-language tasks require bountiful frame inputs that have high computational complexity, but our method efficiently learns near-complete video information.

\subsection{Detailed Ablation Study Results}
\label{sec:ablation}

\begin{table}[H]
    \centering
    
    \resizebox{\linewidth}{!}{%
    \begin{tabular}{l|c|c|c|c|c}
    \toprule
        & \model & w/o Optical Flow & fixed TGB & Uniform Sample & Zero-Shot \\ \midrule
        \textbf{Obj-rel} & 62.27 & 59.13& 62.28 & 53.72& 23.60\\
        \textbf{Rel-act} &51.74 & 15.06 & 47.84 & 48.64 & 17.09\\
        \textbf{Obj-act} &  66.09  & 50.79 & 50.68& 62.10& 29.37 \\
        \textbf{Superlative} &  53.67  & 59.79& 52.12  & 43.84& 28.39\\
        \textbf{Sequencing} & 60.11  & 35.04& 49.43& 55.94 & 48.79\\
        \textbf{Exists} & 60.85 & 60.92 & 60.96 & 55.14  & 48.79\\
        \textbf{Duration} & 36.99  & 26.48 & 40.18 & 40.39 & 26.99 \\
        \textbf{Action} &   0.00 & 0.00& 0.00& 0.28& 0.28 \\
        \textbf{All} & 61.45 & 55.00& 59.88& 54.00 & 25.54   \\ \bottomrule
    \end{tabular}
    }
    \caption{Ablation study of our method on reasoning questions from AGQA 2.0~\cite{GrundeMcLaughlin2021AGQA}.}
    \label{tab:ab_blipagqa2_c}
\end{table}

In ~\cref{tab:ab_blipagqa2_c}, we demonstrate the details of the ablation study of \model on AGQA 2.0. Specifically, we demonstrates the ablation study results of different question types.

\section{Implementation Details}
\label{sec:implem}

\subsection{Details of Datasets} 
\paragraph{Long-form VideoQA.} AGQA is specially designed for compositional spatial-temporal reasoning\footnote{We use AGQA 2.0 which has more balanced distributions.} including 1,455,610/669,207 question answering for train/test splits. NExTQA is a multiple choice \ac{videoqa} benchmark for causal, temporal, and descriptive reasoning, including 52K questions. 
\paragraph{Temporal Question Grounding on Video.} NExT-GQA is an extension of NExT-QA~\cite{Xiao2021NExTQANP} with $10.5K$ temporal grounding labels tied to questions, which contains 3,358/5,553 questions for val/test splits. We report mean Intersection over Union (mIoU), IoU@0.3, and IoU@0.5 as metrics following~\cite{nextgqa}.

\subsection{Implementation Details of \model on Downstream Tasks}
\label{supp:implementation}
The TGB is a 6-layer transformer with RoPE~\cite{su2021roformer}. 
For \model, We use BLIP2-flant5-xl~\cite{li2023blip2} as TGB. For the TGB-augmented framework, we take three vison-language pretraining models as the solver: ALBEF~\cite{Li2021AlignBF}, SINGULARITY~\cite{Lei2022RevealingSF}, and VIOLET~\cite{Fu2021VIOLETE}
For the number of keyframes, we sample 4 frames for \model and 6 frames for TGB-augmented methods to keep consistent with baselines. We take $K=2$ for Gumbel-Softmax tricks in practice. We extract the dense optical flow from the video by RAFT~\cite{2020raft}. For the BLIP2-based model, the total trainable parameters are 195M, thus our framework is lightweight and can be easily adapted to any LLM. All the experiments are performed on NVIDIA A100 80G GPU. 

\subsection{Prompt for Multiple-choice Task on BLIP2}
\label{sec:prompt}

Following ~\cite{yu2023selfchained}, we construct additional prompts to adapt the generative model to the multiple-choice task.
\begin{figure}[H]
    \centering
    \includegraphics[width=\linewidth]{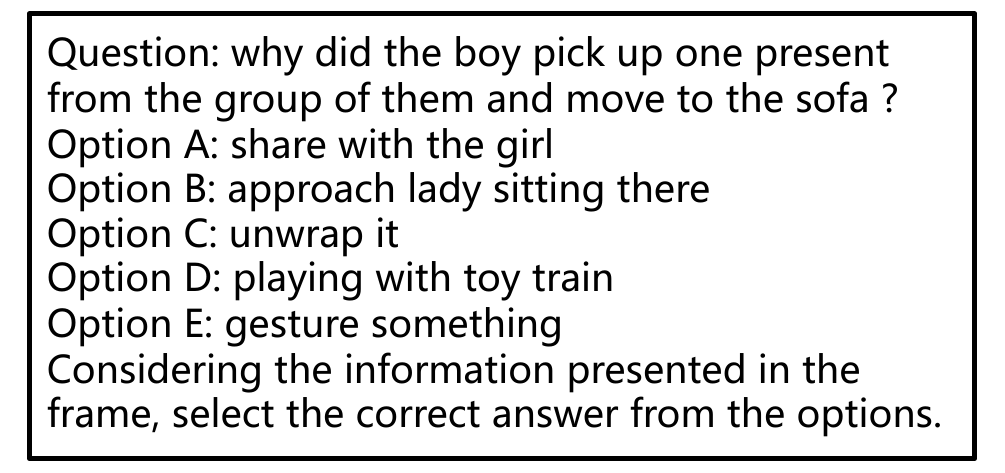}
    \caption{Additional prompt for NExT-MC task}
    \label{fig:prompt}
\end{figure}



\section{Qualitative Studies on AGQA 2.0}
\label{sec:cases}

%
\begin{figure*}[!hbt]
    \centering
    \begin{tabular}{llll}
        
        \rotatebox{90}{\hspace{4mm}Video} & \includegraphics[width=.9\textwidth]{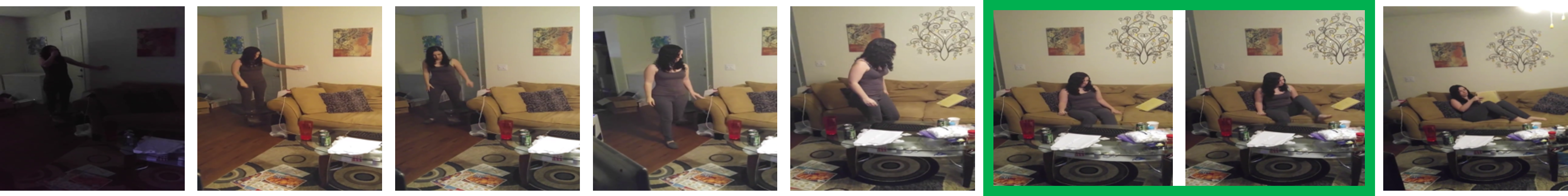} \\
        \rotatebox{90}{\hspace{7mm}OF} & \includegraphics[width=.9\textwidth]{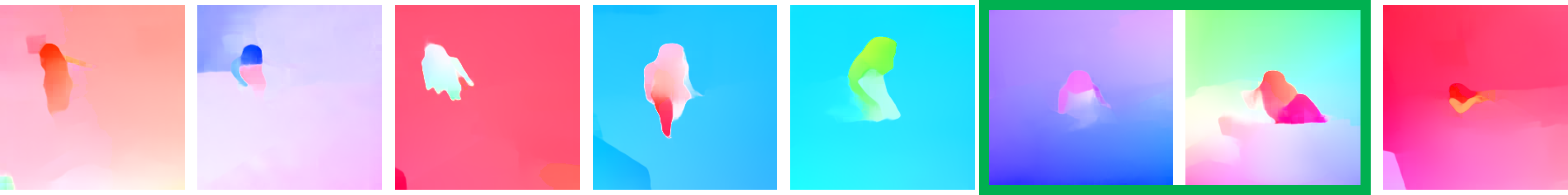} \\
        \multicolumn{2}{p{15cm}}{\textbf{Question:} Before holding a book but after sitting in a bed, what did they undress?} \\
        \multicolumn{3}{p{15cm}}{\textbf{Ground Truth:} shoe \quad{} \textbf{\model:} \textcolor{teal}{\textbf{shoe}} \quad{}  \textbf{BLIP2:} \textcolor{red}{\textbf{dish}} \quad{} \textbf{SEVILA:} \textcolor{red}{\textbf{clothes}}} \\ \\

        \rotatebox{90}{\hspace{4mm}Video} & \includegraphics[width=.9\textwidth]{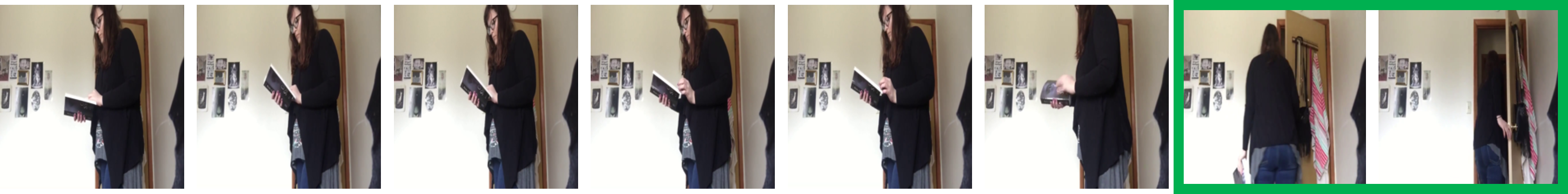} \\
        \rotatebox{90}{\hspace{7mm}OF} & \includegraphics[width=.9\textwidth]{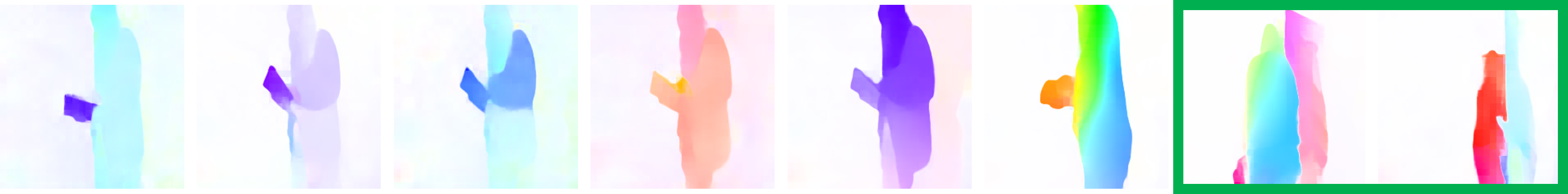} \\
        \multicolumn{2}{p{15cm}}{\textbf{Question:} Which object did the person grasp after watching a book?} \\
        \multicolumn{3}{p{15cm}}{\textbf{Ground Truth:} doorknob \quad{} \textbf{\model:} \textcolor{teal}{\textbf{doorknob}} \quad{} \textbf{BLIP2:} \textcolor{red}{\textbf{NA}} \quad{} \textbf{SEVILA:} \textcolor{red}{\textbf{doorway}}} \\    \\
        
        \end{tabular}
    \caption{Case Studies. OF: Optical Flow. Green and red boxes indicate correct and wrong keyframe predictions, respectively. In these cases, our method could correctly localize the keyframes and predict the right answer. ``NA" indicates the BLIP2 can't generate an answer hitting the answer vocabulary.}
    \label{fig:cases}
\end{figure*}



\begin{figure*}[!t]
    \centering
    \begin{tabular}{ll}
        \rotatebox{90}{\hspace{4mm}Video} & \includegraphics[width=.9\textwidth]{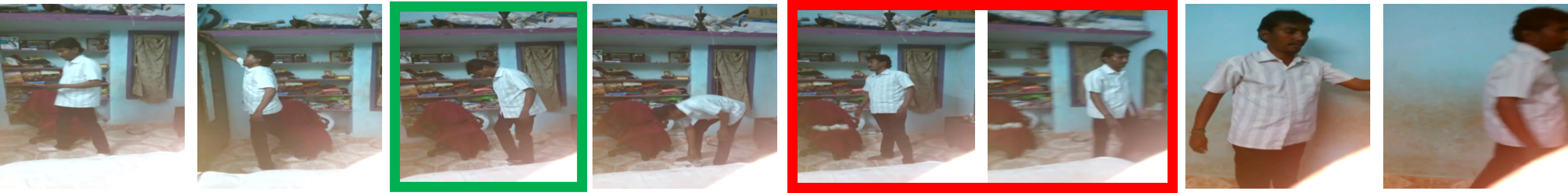} \\
        \rotatebox{90}{\hspace{7mm}OF} & \includegraphics[width=.9\textwidth]{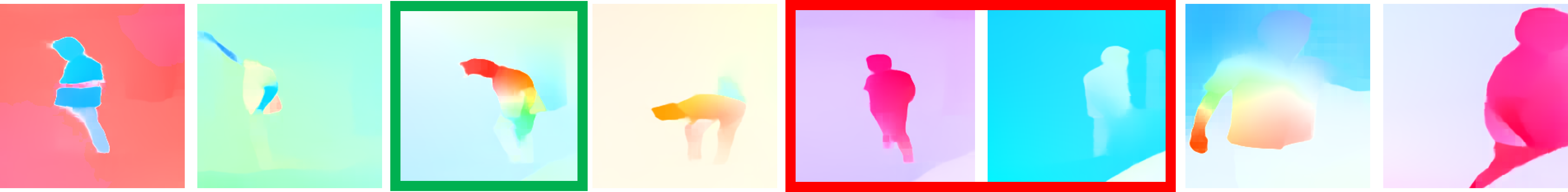} \\
        \multicolumn{2}{p{15cm}}{\textbf{Question:} Between putting a book somewhere and tidying something on the floor, which object were they undressing?} \\
        \multicolumn{2}{p{15cm}}{\textbf{Prediction:} \textcolor{red}{\textbf{shoe}} \quad{} \textbf{Ground Truth:} clothes} \\
        \rotatebox{90}{\hspace{4mm}Video} & \includegraphics[width=.9\textwidth]{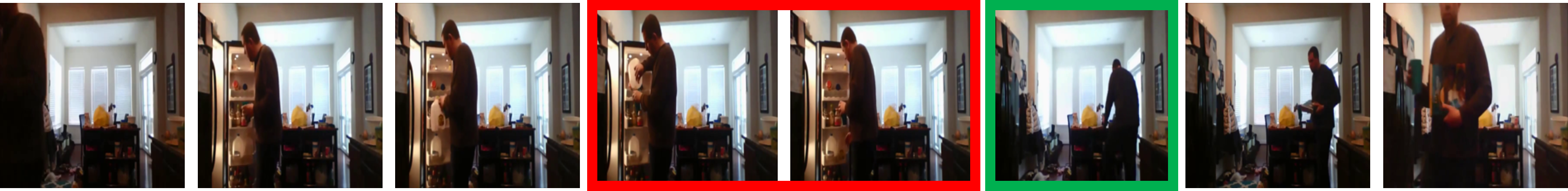} \\
        \rotatebox{90}{\hspace{7mm}OF} & \includegraphics[width=.9\textwidth]{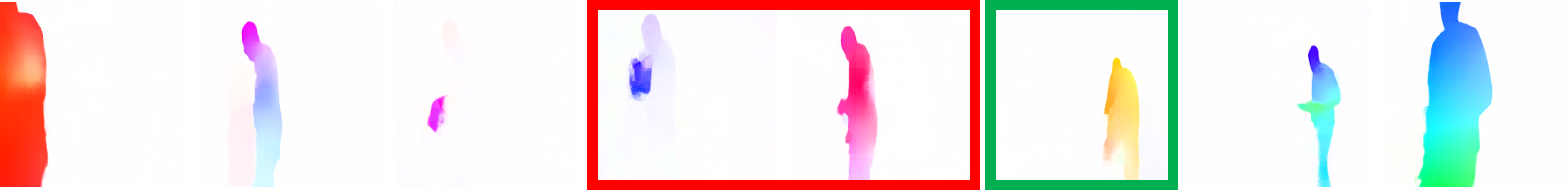} \\
        \multicolumn{2}{p{15cm}}{\textbf{Question:} What was the person taking between putting a cup somewhere and holding a book?} \\
        \multicolumn{2}{p{15cm}}{\textbf{Prediction:} \textcolor{red}{\textbf{box}} \quad{} \textbf{Ground Truth:} food} 
    \end{tabular}
    \caption{Filure Cases. OF: Optical Flow. Green and red boxes indicate correct and wrong keyframe predictions, respectively. For complicated situations involving more than one event, \eg, ``between putting a cup and holding a book", our method could fail to localize the keyframes and thus print the wrong answer.}
    \label{fig:failure-cases}
\end{figure*}

\clearpage
\clearpage

\end{document}